\documentclass[runningheads]{llncs}

\usepackage[colorlinks]{hyperref}
\usepackage{url}
\usepackage{amsmath}
\usepackage{algorithm}
\usepackage{algorithmic}

\usepackage{lipsum}
\usepackage{epsf}
\usepackage{multirow}
\usepackage{booktabs}
\usepackage[dvips]{graphicx}
\usepackage{subfigure}
\usepackage[normalem]{ulem}
\usepackage{amsmath}
\usepackage{amssymb}
\usepackage{mathrsfs}
\usepackage{float}
\usepackage[colorinlistoftodos]{todonotes}
\usepackage[misc]{ifsym}

\newcommand{\name}{CSMVH}

\begin{document}
%
%\title{Contribution Title\thanks{Supported by organization x.}}
\title{Central Similarity Multi-View Hashing for Multimedia Retrieval}
\titlerunning{\name}
\author{Jian Zhu\inst{1\dagger} \and
Wen Cheng\inst{1\dagger} \and
Yu Cui\inst{1\dagger}\and
Chang Tang\inst{2}\and
Yuyang Dai\inst{1}\and
Yong Li\inst{1}\and
Lingfang Zeng\inst{1*}}
\authorrunning{F. Author et al.}
\institute{Zhejiang Lab \\ \email{\{qijian.zhu, chengwen, yu.cui, daiyy, yonglii, zenglf\}@zhejianglab.com} \and
China Uniersity of Geosciences \\
\email{tangchang@cug.edu.cn}\\}
% \author{Jian Zhu \and Wen Cheng}
\toctitle{CSMVH}
\tocauthor{}
\authorrunning{Jian Zhu et al.}
% \institute{}
%
%\maketitle              % typeset the header of the contribution

%

% \maketitle              % typeset the header of the contribution
% %
\maketitle
\def\thefootnote{$\dagger$}\footnotetext{These authors contributed equally to this work.}
\def\thefootnote{*}\footnotetext{Corresponding author.}
\begin{abstract}
Hash representation learning of multi-view heterogeneous data is the key to improving the accuracy of multimedia retrieval. However, existing methods utilize local similarity and fall short of deeply fusing the multi-view features, resulting in poor retrieval accuracy. Current methods only use local similarity to train their model. These methods ignore global similarity. Furthermore, most recent works fuse the multi-view features via a weighted sum or concatenation. We contend that these fusion methods  are insufficient for capturing the interaction between various views. We present a novel \underline{C}entral \underline{S}imilarity \underline{M}ulti-\underline{V}iew \underline{H}ashing (CSMVH) method to address the mentioned problems. Central similarity learning is used for solving the local similarity problem, which can utilize the global similarity between the hash center and samples. We present copious empirical data demonstrating the superiority of gate-based fusion over conventional approaches. On the MS COCO and NUS-WIDE, the proposed CSMVH performs better than the state-of-the-art methods by a large margin (up to $11.41$ mean Average Precision (mAP) improvement).
\end{abstract}
\section{Introduction}
Multi-view hashing solves multimedia retrieval problems. The accuracy can be significantly increased with a well-crafted multi-view hashing method. Multi-view hashing, as opposed to single-view hashing, which only searches in a single view, can make use of data from other sources (e.g., image, text, audio, and video). Extraction of heterogeneous features from several views is accomplished via multi-view hashing representation learning. It fuses multi-view features to capture the complementarity of different views.

\begin{figure}[htp]
	\centering
	\includegraphics[width=10cm]{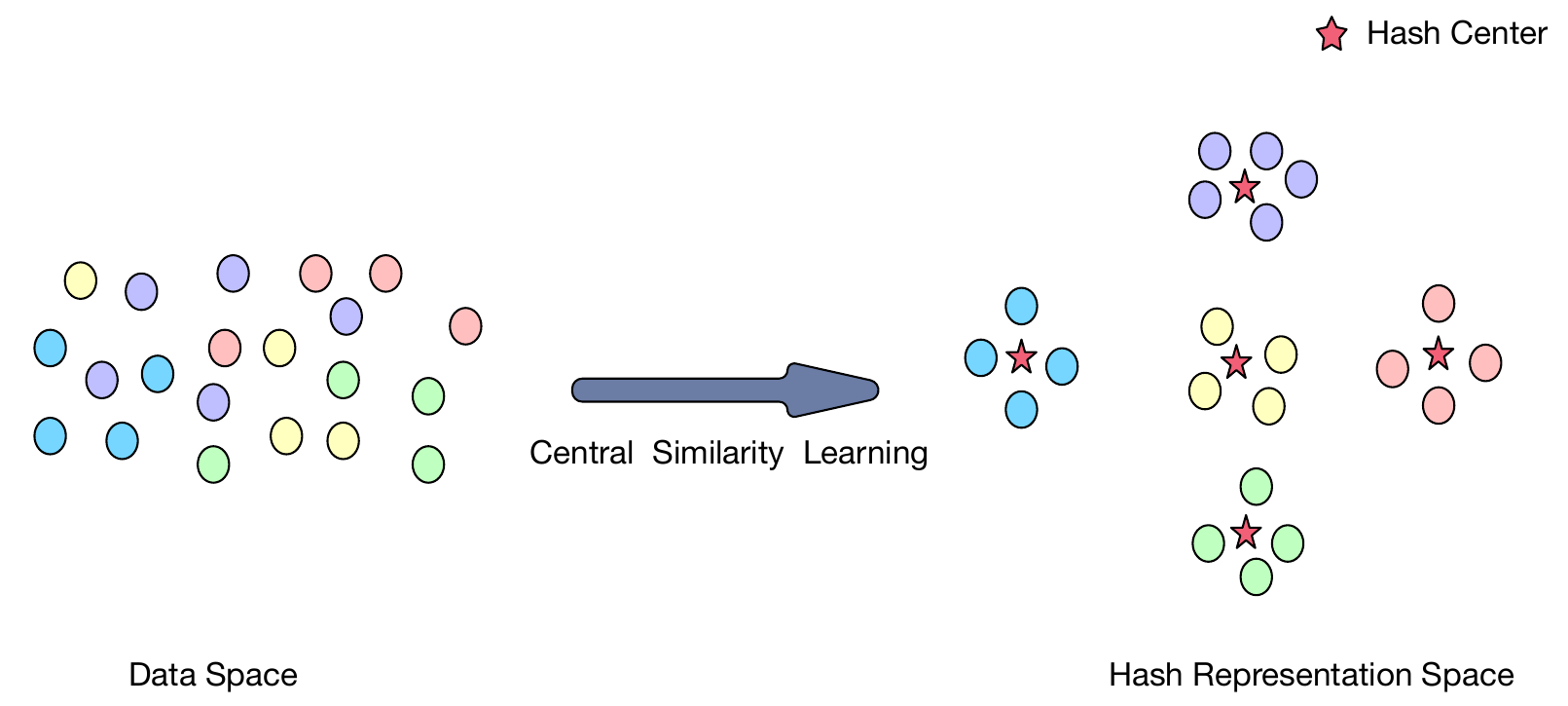}
	\caption{The inputs are randomly distributed in the data space. Central similarity learning utilizes the hash center to separate samples from different classes while minimizing the class intra-distance.}
	\label{fig:01}
\end{figure}

Retrieval accuracy is currently poor for multi-view hashing methods. The reasons are as follows. First, current methods are fascinated by the information provided by the local similarity. The global similarity is not given enough credit. For example, Flexible Graph Convolutional Multi-modal Hashing (FGCMH) \cite{lu:18} is a GCN-based \cite{welling:50} multi-view hashing method. It constructs the edges of the graph based on similarity. The GCN then aggregates features of adjacent nodes. Hence, the global similarity is not taken into account during this approach. Second, it is not enough to fuse multiple views. To get a global embedding, common multi-view hashing methods (e.g., Deep Collaborative Multi-View Hashing (DCMVH) \cite{zhu:17}, Flexible Multi-modal Hashing (FMH) \cite{zhu:52}) use weighted sum or concatenation to fuse the features. During the fusing process, the relationship between the text and image is disregarded, which results in a lack of expressiveness in the fused features.
%Finally,  the expressiveness of the backbone networks is insufficient. For instance, Fast Discrete Collaborative Multi-modal Hashing  (FDCMH) \cite{zheng:57} and  Online Multi-modal Hashing with Dynamic Query-adaption (OMH-DQ) \cite{lu:60} hire a VGG net \cite{simonyan:51} for the image modal and a Bag-of-Words (BoW) model \cite{zhang:53} for the text view. These methods are outdated for feature extraction. Thus, an update in feature extraction methods is necessary. 
The aforementioned facts result in low overall retrieval accuracy. 

This paper proposes \textit{Central Similarity Multi-View Hashing} method termed CSMVH, which introduces central similarity learning to multi-view hashing. As shown in Fig. \ref{fig:01}, samples are distributed randomly in the raw data space. Using central similarity learning, semantically similar samples are close to one another, while semantically different samples are pushed away from each other. This can improve the ability to distinguish sample semantic space. At the same time, because central similarity learning also has the advantage of linear complexity, it is a very effective method. 

CSMVH takes advantage of the Gated Multimodal Unit (GMU) \cite{arevalo:38} to learn the interaction and dependency between the image and text. Different from typical methods, our method fuses multi-view features into a global representation without losing dependency. The semantics-preservation principle of hash representation learning governs the selection of the best embedding space.

%CSMVH only has a time complexity of $O(n)$, where $n$ is the number of samples. To achieve similar results, typical methods utilize pairwise or triplet data similarity, which has a time complexity of $O(n^{2})$ or $O(n^{3})$. Therefore, CSMVH is more efficient than previous methods. Furthermore, CSMVH also performs well on imbalanced datasets thanks to the promising initialization of central similarity learning. 

We evaluate our method CSMVH on the multi-view hash representation learning tasks of MS COCO and NUS-WIDE datasets. The proposed method provides up to $11.41\%$ mAP improvement in benchmarks. Our main contributions are as follows:
\begin{itemize}
    \item We propose a novel multi-view hash method CSMVH. The proposed method achieves state-of-the-art results in multimedia retrieval.
    \item Central similarity learning is introduced to multi-view hashing for the first time. Our method has lower time complexity than typical pairwise or triplet similarity methods. And it converges faster than previous methods.
    \item We take advantage of the GMU to learn a better global representation of different views to address the insufficient fusion problem. 
\end{itemize}

\section{Related Work}
Multi-view hashing \cite{zhu2023deep,liu:7,kang:8,song:9,chen2023optimization,xu2022towards,liu:10,shen:11,yang:12,kim:13} fuses multi-view features for hash representation learning. These methods use a graph to model the relationships among different views for hash representation learning. Multiple Feature Hashing (MFH) \cite{song:9} not only preserves the local structure information of each view but also considers global information during the optimization. Multi-view Alignment Hashing (MAH) \cite{liu:10} focuses on the hidden semantic information and captures the joint distribution of the data. Multi-view Discrete Hashing (MvDH) \cite{shen:11} performs spectral clustering to learn cluster labels. Multi-view Latent Hashing (MVLH) \cite{shen:6} learns shared multi-view hash codes from a unified kernel feature space.  Compact Kernel Hashing with Multiple Features (MFKH) \cite{liu:7} treats supervised multi-view hash representation learning as a similarity preserving problem. Different from MFKH, Discrete Multi-view Hashing (DMVH) \cite{yang:12} constructs a similarity graph based on Locally Linear Embedding (LLE)  \cite{hou:14,saul:15}. 

Recently, some deep learning-based multi-view hashing methods are proposed. Flexible Discrete Multi-view Hashing (FDMH) \cite{liu:23}, Flexible Online Multi-modal Hashing (FOMH) \cite{lu:24}, and Supervised Adaptive Partial Multi-view Hashing (SAPMH) \cite{zheng:25} seek for a projection from input space to embedding space using nonlinear methods. The learned embeddings are fused into multi-modal embedding for multi-view hashing. Instead of seeking an embedding space, Deep Collaborative Multi-view Hashing (DCMVH) \cite{zhu:17} directly learns hash codes using a deep architecture. A discriminative dual-level semantic method is used for their supervised training. FGCMH \cite{lu:18} is based on a graph convolutional network (GCN). It preserves both the modality-individual and modality-fused structural similarity for hash representation learning. To facilitate multi-view hash at the concept aspect, Bit-aware Semantic Transformer Hashing (BSTH) \cite{tan:61} explores bit-wise semantic concepts while aligning disparate modalities.

Compared with previous multi-view hashing methods, we use central similarity learning and gate-based fusion for the first time. First, current multi-view methods underrate the importance of global similarity and only obtain information provided by local similarity. Therefore, we introduce central similarity learning, which can take advantage of global similarity to make our method learn more helpful information. Second, current hashing methods fuse multi-view features insufficiently, leading to a weak expressiveness of the obtained global representation. To address this issue, we adopt gate-based fusion to learn the interaction and dependency between the image and text features.

\section{The Proposed Methodology}
The goal of CSMVH is to use central similarity learning to train a deep multi-view hashing network. We first propose the deep multi-view hashing network, which uses gate-based fusion for the multi-view features. Then the loss of central similarity learning is turned to illustrate. Eventually, CSMVH reduces the computational complexity. 

\subsection{Deep Multi-View Hashing Network}
\begin{figure*}[htp]
	\centering
	\includegraphics[width=11.72cm]{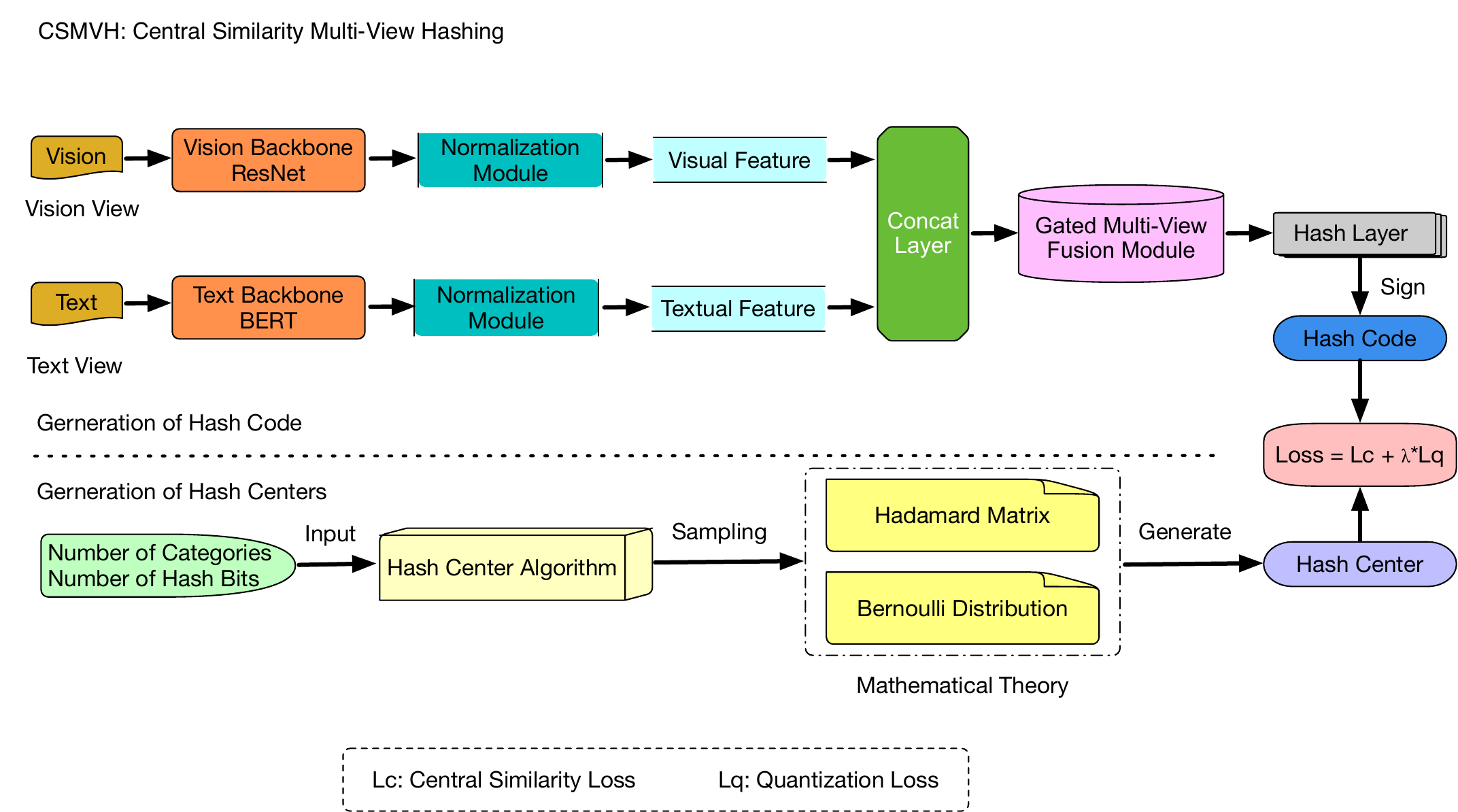}
	\caption{Architecture Overview of CSMVH. CSMVH has two pipelines, one for the multi-view hash code generation and the other for the hash center generation. In the first pipeline, the image and text features are extracted by the backbones, and then the features are fused by the GMU and hashed by the hash layer. In the second pipeline, the hash center for each class is generated using the Hadamard matrix or Bernoulli distribution. Eventually, CSMVH is optimized by minimizing the distance between the hash code and the corresponding hash center while considering the quantization loss.}
	\label{fig:02}
\end{figure*}

The deep multi-view hashing network is made to transform multi-view data into hash code. As shown in Fig. \ref{fig:02}, CSMVH comprises a vision backbone, text backbone, normalization modules, multi-view fusion module, and a hash layer. We utilize deep residual net (ResNet) \cite{he:5}  for image feature extraction and BERT-base \cite{devlin:3}  for text feature extraction. The extracted features are then passed to the fusion module. These modules are described in detail below. In order to better explain our method, we detailly write the steps of the CSMVH in Algorithm \ref{alg:algorithm}.

\begin{algorithm}[htbp]
  \caption{Central Similarity Multi-View Hashing}
  \label{alg:algorithm}
  \textbf{Input}: Training Set $\left\{\mathbf{x_i}^{(m)}, \mathbf{Y_i}\right\}_{i=1}^{N}$, Number of views $m$, Number of training sets $N$, Hash Code Length $K$, Hash Centers $\mathcal{HC}=\left\{hc_{i}\right\}_{i=1}^{V} \subset\{-1,1\}^{K}$, $V$ is the number of categories of $Y$. Hyper-parameter $\lambda$, Training epochs $T$.
  \begin{algorithmic}[1] %[1] enables line numbers
    \STATE Initialize BERT parameters $w_{BERT}$ and ResNet parameters $w_{ResNet}$ by loading pre-trained model parameters.
    \STATE Randomly initialize Normalization Module parameters $w_{vnorm}$ and $w_{tnorm}$, Multi-View Fusion Module parameters $w_{fusion}$ and Hash Layer parameters $w_{hash}$.
    \STATE Initialize learning rate, mini-batch size, dropout, and $\lambda$.
    \FOR{$epoch=1$ to $T$}
    \FOR{$i=1$ to $N$} 
    \STATE Backbones extract embedding from view data.
    \STATE Normalize embedding to the same dimensions.
    \STATE Fuse multi-view embedding to the global representation.
    \STATE Hash Layer generates hash code.
    \STATE Update $L_{total}$ by Eq.\eqref{eq:loss}.
    \STATE Update $w_{vnorm}$, $w_{tnorm}$, $w_{fusion}$, and $w_{hash}$ by the Adam Optimizer.
    
    \ENDFOR
    \ENDFOR
  \end{algorithmic}
  \textbf{Output}: Multi-View Hash Network Parameters: $\mathbf{W}_{K}^{h a s h}$.
  
\end{algorithm}

\subsubsection{Vision Encoder}
We use a deep residual network \cite{he:5} to extract features. Specifically, we take ResNet-50 as the backbone network for acquiring visual features.

\subsubsection{Text Encoder}
We use BERT-base \cite{devlin:3} model as the backbone network for extracting text features. We transform the semantic tag of images into descriptive sentences and then input it to the BERT-base model for feature extraction. Finally, the CLS Token embedding of the BERT model output layer is taken as the semantic vector of the image tag.

\subsubsection{Normalization Module}
The normalization module projects multi-view features into the same dimension, and the output range of features is also within a certain threshold. It is implemented through a fully connected layer network.

\subsubsection{Gated Multi-View Fusion Module}
In multimodal hashing, we need to fuse different information sources so that the fused one can provide more information than a single source. Fusion can be performed at the feature level or decision level. Gate mechanism can combine both feature and decision fusion. In this work, we utilize a gate-based fusion module called GMU. GMU is inspired by GRU \cite{chung:58} and LSTM \cite{hochreiter:59}. It can work as an internal unit finding intermediate representation for multi-view data. The data from different sources is modulated. The gate will learn a prediction from input data to fusion weight. The modulated data is combined using weighted summation. Primarily, for image and text input, the equation governing can be represented as follows:
\begin{equation}
  h_{i}=\tanh \left(W_{i} \cdot x_{i}\right)
\end{equation}

\begin{equation}
  h_{t}=\tanh \left(W_{t} \cdot x_{t}\right)
\end{equation}

\begin{equation}
  z=\sigma\left(W_{z} \cdot\left[x_{i}, x_{t}\right]\right)
\end{equation}

\begin{equation}
  h_{f}=z * h_{i}+(1-z) * h_{t}
\end{equation}

Where $x_{i}$ and $x_{t}$ are the image and text input, $h_{i}$  and $h_{t}$ are the modulated feature. $\sigma$ represents the sigmoid activation function. $z$ is the predicted weight for summation. We let $\Theta=\left\{W_{i}, W_{t}, W_{z}\right\}$, where $\Theta$ is the learnable parameters. $h_{f}$  is the global vector after multi-view fusion.

\subsubsection{Hash Layer} 
The hash layer is a linear layer with a $tanh$ activation, which is denoted by:
\begin{equation}
h_{\text{k-bit}} = \text{sgn}[\tanh(w_{\text{hash}}X_{\text{fusion}}+b_{\text{hash}})],
\end{equation}
where $sgn$ is the signum function. $w_{\text{hash}} \in \mathbb{R}^{n \times n}$  and $b_{\text{hash}} \in \mathbb{R}^{n}$ are network parameters. The output has exactly as many dimensions as the hash code has. Each dimension corresponds to a hash bit.

\subsection{Central Similarity Learning}
Let $\mathcal{X}= \left\{\left\{{x_{i}}^{(m)}\right\}, Y_i\right\}_{i=1}^{N}$ denotes the train dataset, where each ${x_{i}}^{(m)} \in \mathbb{R}^{D}$ is a multi-view instance, $m$ is the number of views, and $N$ is the total number of dataset samples. And $Y = \left\{y_{1}, y_{2}, \ldots, y_{N}\right\}$ is set, where $y_{i}$ represents the label of $x_{i}$. We use the symbol $F: x \mapsto he \in\{-1,1\}^{K}$ to signify the deep hash function from the input space $\mathbb{R}^{D}$ to K-bit Hamming space $\{-1,1\}^{K}$. Like other deep hashing methods, we investigate a deep multi-view hashing method, which can generate hash code approximation for data points with the same semantic label in Hamming space.

The use of central similarity learning can cause data points from the same class to cluster together around a single hash center and those from other semantic categories to cluster together around several hash centers, respectively. It makes sense that hash centers maintain a suitable mutual separation from one another, which effectively keeps samples from various classes apart in the Hamming space. High-quality hash codes can be produced through central similarity learning in the deep hash function $F$ by learning the overall similarity information between data pairs.

\subsubsection{Hash Center Definition}
We use a similar definition of hash centers as Central Similarity Quantization (CSQ) \cite{yuan:1} with a useful modification. We replace the hash code 0 with -1, thus    $\mathcal{HC}=\left\{hc_{i}\right\}_{i=1}^{V} \subset\{-1,1\}^{K}$.  $V$ is the number of categories of $Y$. An elegant linear relationship exists between Hamming distance  $\operatorname{dist}_H(\cdot, \cdot)$  and inner product $\langle\cdot, \cdot\rangle$: 
\begin{equation}
{dist}_H\left(hc_i, hc_j\right)=\frac{1}{2}\left(K-\left\langle hc_i, hc_j\right\rangle\right),
\end{equation}
where $K$ is the length of hash bits. We can easily have the following:
\begin{equation}
    \frac{1}{\mathrm{~T}} \sum_{i \neq j}^V\left\langle h c_i, h c_j\right\rangle \leq 0
\end{equation}
$V$ is the number of hash centers, and T is the number of combinations of different $hc_{i}$ and $hc_{j} \in \mathcal{HC}$.

\subsubsection{Generation of Hash Centers}
With the definition of hash center, we can naturally notice a boundary scenario:
\begin{equation}
    \frac{1}{\mathrm{~T}} \sum_{i \neq j}^V\left\langle h c_i, h c_j\right\rangle=0
\end{equation}
If all hash centers are orthogonal to each other, this equation holds. The most intuitive way to generate orthogonal vectors is by leveraging the Hadamard matrix. The rows of the Hadamard matrix are orthogonal to each other. Thus, we can easily get the hash center with the Hadamard matrix and concatenated transpose. However, we need Bernoulli Distributions for the extra when we need more than $2K$ hash centers.

\subsubsection{Loss Function}
Given the generated centers $\mathcal{HC}=\left\{hc_{i}\right\}_{i=1}^{V} \subset\{-1,1\}^{K}$ in the K-dimensional Hamming space for training data $X$ with $V$ categories, we obtain the semantic hash centers $\mathcal{HC}^{\prime}=\left\{hc_{1}^{\prime}, hc_{2}^{\prime}, \ldots, hc_{N}^{\prime}\right\}$ for single-label or multi-label data, where $hc_{i}^{\prime}$ denotes the hash center of the data sample $x_{i}$. 
Since hash centers are binary vectors, we use Binary Cross Entropy (BCE) to measure the Hamming distance between the hash code and its center, $dist_{H}\left(hc_{i}^{\prime}, he_{i}\right)={BCE}\left(hc_{i}^{\prime}, he_{i}\right)$. We obtain the optimization objective of the central similarity loss $L_{c}$:
\begin{equation}
    hc_{i, k}^{\prime} =\frac{1}{2} \left(1+hc_{i, k}^{\prime}\right) 
\end{equation}

\begin{equation}
    he_{i, k} =\frac{1}{2} \left(1+he_{i, k}\right) 
\end{equation}

\begin{footnotesize}
    \begin{equation}
        L_{c}= \frac{1}{N}  \sum_{i=1}^{N}  \sum_{k=1}^ {K}\left[hc_{i, k}^{\prime} \log he_{i, k}+\left(1-hc_{i, k}^{\prime}\right) \log \left(1-he_{i, k}\right)\right]
    \end{equation}
\end{footnotesize}
The output of the neural network is continuous. However, in a Multi-view hashing problem, the output is binary code. Hence, an intuitive method uses a threshold to quantify the network output. Unfortunately, the foundation of the elegant linear relation between Hamming distance and inner product requires the outputs to be exact -1 and 1. Otherwise, it does not hold. Furthermore, quantifying the output using a threshold will introduce an uncontrollable quantization error, which can not pass through the backpropagation. Inspired by DHN \cite{zhu:37}, we utilize a quantization before constraining the network output. Different from DHN, we use the log cosh function rather than L1-norm for its smoothness, which can be represented as:
\begin{equation}
    L_{q}=\frac{1}{N} \sum_{i=1}^{N} \sum_{k=1}^{K}\left(\log \cosh \left(\left|he_{i, k}\right|-1\right)\right.
\end{equation}
Eventually, we have a central similarity optimization problem:
\begin{equation}
\label{eq:loss}
    L_{total}=L_{c}+\lambda L_{q}
\end{equation}
where $L_{total}$ is the total loss function of our algorithm. $\lambda$ is the hyper-parameter obtained through grid search in our work.

\subsubsection{Computational Complexity}
Concerning time complexity, CSMVH only has a $O(n)$ value, where $n$ is the number of samples. To achieve similar results, most approaches use pairwise or triplet data similarity, which has a time complexity of $O(n^{2})$ or $O(n^{3})$. As a result, CSMVH is more effective than the earlier method. For example, Flexible Graph Convolutional Multi-modal Hashing (FGCMH) \cite{lu:18} is a GCN-based multi-view hashing method. It requires an adjacency matrix to represent the relationship between the nodes. Therefore, the complexity of FGCMH is $O(n^2)$. In addition, Eq. \eqref{eq:loss} can also be concluded that the complexity of our method is $O(n)$. 

Through the above analysis, we use central similarity learning to improve the computational complexity, which is that our CSMVH converges faster than the previous methods.

\section{Experiments}
We evaluate the proposed CSMVH on large-scale multimedia retrieval tasks in experiments. Two genetic datasets are adopted: NUS-WIDE \cite{chua:22} and MS COCO \cite{lin:21}. These datasets have been widely used for evaluating multimedia retrieval performance. We use the mean Average Precision (mAP) as the evaluation metric. The statistics of two datasets used in experiments are summarized in Table \ref{Tab:01}.

\subsection{Evaluation Datasets}
\begin{table*}[]
    \centering
     \caption{General statistics of two datasets. The dataset size, number of categories, and feature dimensions are included.}
    \resizebox{\textwidth}{!}{\begin{tabular}{llllllll}
        \toprule[1pt]
        Datasets   & Training Size & Retrieval Size & Query Size & Categories&Visual Feature & Text Feature \\ \midrule[0.8pt]
        MS COCO & 18000  & 82783 & 5981    & 80&ResNet(768-D) &BERT(768-D)\\
        NUS-WIDE & 21000  & 193749 & 2085    & 21&ResNet(768-D) &BERT(768-D)\\
        \bottomrule[1pt]
    \end{tabular}}
   
    \label{Tab:01}
\end{table*}

\subsubsection{MS COCO} 
In our experiments, the MS COCO 2014 dataset is adopted. It contains 82,783 training samples and 40,504 validation samples. We randomly select 80 validation samples from each category as the query set to retrieve samples from the training set and use 18000 of them for training.

\subsubsection{NUS-WIDE}
NUS-WIDE contains 269,648 Flickr images with 81 ground-truth semantic concepts. In experiments, we select the 21 most common concepts. We randomly select 100 samples for each concept as the query set. The rest of the images are treated as the retrieval set. We utilize 21,000 samples from the retrieval set for training. 

\subsection{Evaluation Metrics}
To evaluate the metric of multi-view hashing methods, mean Average Precision (mAP) is utilized. The Hamming ranking measure can be assessed with good steady using the mAP. It is described as follows:

\begin{equation}
  A P(q)=\frac{1}{M} \sum_{r=1}^{R} P_{q}(r) \text{Ind(r)}
\end{equation}

\begin{equation}
  m A P=\frac{1}{Q} \sum_{i=1}^{Q} A P\left(q_{i}\right)
\end{equation}

where $Q$ is the number of queries and $P_{q}$ is the precision
for query $q$ when the top $r^{t h}$ similar searching results returned, $Ind(r)$ is an
indicator function which is 1 when the $r^{th}$ result has the same label with $q$ and otherwise 0, $M$ is the number of same label samples of query $q$, and $R$ is the size of the retrieval dataset.

\subsection{Implementation Details}
Our code is implemented on the PyTorch platform. A ResNet-50 pre-trained on ImageNet is employed as
the backbone. We use the BERT-base model as the text backbone. The image and text feature outputs are set to 768-dimensional by normalization modules. The dropout probability is set to 0.1 to improve the generalization capability. The combination coefficient $\lambda$ of the total loss function is $0.25$.

\subsection{Baseline}
To evaluate the retrieval metric, the proposed CSMVH is compared with twelve comparable multi-view hashing methods, including four unsupervised methods (e.g., Multiple Feature Hashing (MFH) \cite{song:9}, Multi-view Alignment Hashing (MAH) \cite{liu:10}, Multi-view Latent Hashing (MVLH) \cite{shen:6}, and Multi-view Discrete Hashing (MvDH) \cite{shen:11}) and eight supervised methods (e.g., Multiple Feature Kernel Hashing (MFKH) \cite{liu:7}, Discrete Multi-view Hashing (DMVH) \cite{yang:12}, Flexible Discrete Multi-view Hashing (FDMH) \cite{liu:23}, Flexible Online Multi-modal Hashing (FOMH) \cite{lu:24}, Deep Collaborative Multi-View Hashing (DCMVH) \cite{zhu:17}, Supervised Adaptive Partial Multi-view Hashing (SAPMH) \cite{zheng:25}, Flexible Graph Convolutional Multi-modal Hashing (FGCMH) \cite{lu:18}, and Bit-aware Semantic Transformer Hashing (BSTH) \cite{tan:61}).

\subsection{Analysis of Experimental Results}

\begin{table*}
    \setlength{\tabcolsep}{2pt}
    \centering
    \caption{The comparable mAP results on NUS-WIDE and MS COCO. The experimental conditions of all methods are the same. The best results are bolded, and the second-best results are underlined. The * indicates that the results of our method on this dataset are statistical significance.}
    \resizebox{\textwidth}{!}{\begin{tabular}{llllllllllllll}
        \toprule[1pt]
        \multicolumn{1}{c}{\multirow{2}{*}{Methods}} & \multicolumn{1}{c}{\multirow{2}{*}{Ref.}}     & \multicolumn{4}{c}{NUS-WIDE*}      & \multicolumn{4}{c}{MS   COCO*}       \\  \cmidrule(r){3-6}  \cmidrule(r){7-10}  
        \multicolumn{1}{c}{}                         & \multicolumn{1}{c}{}                      & 16 bits & 32 bits & 64 bits & 128 bits & 16 bits & 32 bits & 64 bits & 128 bits \\ \midrule[0.8pt]
        MFH                                          & TMM13                                     & 0.3603 & 0.3611 & 0.3625 & 0.3629  & 0.3948 & 0.3699 & 0.3960  & 0.3980   \\
        MAH                                          & TIP15                                       & 0.4633 & 0.4945 & 0.5381 & 0.5476  & 0.3967 & 0.3943 & 0.3966 & 0.3988  \\
        MVLH                                         & MM15                                        & 0.4182 & 0.4092 & 0.3789 & 0.3897  & 0.3993 & 0.4012 & 0.4065 & 0.4099  \\
        MvDH                                         & TIST18                                     & 0.4947 & 0.5661 & 0.5789 & 0.6122  & 0.3978 & 0.3966 & 0.3977 & 0.3998  \\ \midrule[0.8pt]
        MFKH                                         & MM12                                      & 0.4768 & 0.4359 & 0.4342 & 0.3956  & 0.4216 & 0.4211 & 0.4230  & 0.4229  \\
        DMVH                                         & ICMR17                                     & 0.5676 & 0.5883 & 0.6902 & 0.6279  & 0.4123 & 0.4288 & 0.4355 & 0.4563  \\
        FOMH                                         & MM19                                       & 0.6329 & 0.6456 & 0.6678 & 0.6791  & 0.5008 & 0.5148 & 0.5172 & 0.5294  \\
        FDMH                                         & NPL20                                       & 0.6575 & 0.6665 & 0.6712 & 0.6823  & 0.5404 & 0.5485 & 0.5600   & 0.5674  \\
        DCMVH                                        & TIP20                                      & 0.6509 & 0.6625 & 0.6905 & 0.7023  & 0.5387 & 0.5427 & 0.5490  & 0.5576  \\
        SAPMH                                        & TMM21                                        & 0.6503 & 0.6703 & 0.6898 & 0.6901  & 0.5467 & 0.5502 & 0.5563 & 0.5672  \\
        FGCMH                                        & MM21                                       & 0.6677 & 0.6874 & 0.6936 & 0.7011  & 0.5641 & 0.5273 & 0.5797 & 0.5862  \\
  BSTH                                        & SIGIR22                                       & \underline{0.6990} & \underline{0.7340} & \underline{0.7505} & \underline{0.7704}  & \underline{0.5831} & \underline{0.6245} & \underline{0.6459} & \underline{0.6654}  \\\midrule[0.8pt]
        CSMVH     & Proposed   & \textbf{0.7360} & \textbf{0.7633} & \textbf{0.7740} & \textbf{0.7819} & \textbf{0.6028} & \textbf{0.7002}  & \textbf{0.7485} & \textbf{0.7795} \\
        \bottomrule[1pt]
    \end{tabular}}
    
    \label{Tab:02}
\end{table*}

\begin{table*}[!t]
    \setlength{\tabcolsep}{2pt}
    \centering
    \caption{Ablation experiments on two datasets. The mAP is shown in the table, indicating the performance impact of different modules.}
    \resizebox{\textwidth}{!}{\begin{tabular}{lllllllllllll}
        \toprule[1pt]
        \multicolumn{1}{c}{\multirow{2}{*}{Methods}} &  \multicolumn{4}{c}{NUS-WIDE}  & \multicolumn{4}{c}{MS COCO} \\   \cmidrule(r){2-5}  \cmidrule(r){6-9}  \cmidrule(r){10-13}
        \multicolumn{1}{c}{} & 16 bits & 32 bits & 64 bits & 128 bits& 16 bits & 32 bits & 64 bits & 128 bits \\  \midrule[0.8pt] 
        CSMVH-central     & 0.7352 & 0.7594 & 0.7683& 0.7798& 0.5892& 0.6893&0.7361&0.7731\\
        CSMVH-quant      & 0.3085 & 0.3085 & 0.3085 & 0.3085 &0.3502&0.3502&0.3502&0.3502\\
        CSMVH-text     & 0.4873 & 0.5176 & 0.5196 & 0.5242 &0.5548&0.6134&0.6407&0.6784\\
        CSMVH-image     & 0.7208& 0.7512& 0.7623 & 0.7679 & 0.5459&0.6454&0.6833&0.7197 \\
        CSMVH-concat     & 0.7283 & 0.7566 & 0.7647 & 0.7718   & 0.5844&0.6864&0.7282&0.7580 \\ \midrule[0.8pt]
        CSMVH    & \textbf{0.7360} & \textbf{0.7633} & \textbf{0.7740} & \textbf{0.7819} & \textbf{0.6028} & \textbf{0.7002}  & \textbf{0.7485} & \textbf{0.7795} \\
        \bottomrule[1pt]
    \end{tabular}}
    
    \label{Tab:03}
\end{table*}
The mAP results are presented in Table \ref{Tab:02}. The experimental comparisons of all methods are conducted according to the unified conditions of the train set, the retrieval set, and the query set in Table \ref{Tab:01}. The proposed CSMVH is a top performer in multi-view hashing tasks. On the NUS-WIDE dataset, our method outperforms the prior state-of-the-art method by a large margin (up to 3.70\%). However, the more complex dataset is where our method shines. MS COCO contains 80 categories, and the samples are more complex than other datasets. Our method beats the previous state-of-the-art 128-bit hashing result with only a 16-bit hash code. Our method shows a magnificent retrieval accuracy improvement with the number of hash bits increasing. The mAP of BSTH \cite{tan:61} only increased 8.23\% from 16-bit to 128-bit, while CSMVH has a 17.67\% absolute mAP increase, which indicates a great capability of solving very complex hashing problems. The absolute 128-bit mAP increase over BSTH is 11.47\% on the MS COCO dataset. 

The main reasons for these superior results come from two aspects: 
\begin{itemize}
\item Central similarity learning can take advantage of global similarity to make CSMVH learn more useful information. It enhances the discriminative and semantic capability of hash codes. This can improve the accuracy of multimedia retrieval.
\item The multi-view fusion module could deeply fuse the multi-view features into a global representation. It can fully explore the complementarity of multi-view features and improve the ability of feature expression. This can generate high-quality hash codes.
\end{itemize}

\subsection{Ablation Studies}
To evaluate the proposed CSMVH component by component, we perform ablation with different experiment settings of our method and report the performance. The experiment settings are as follows:
\begin{itemize}
    \item \emph{CSMVH-central}: The BCE loss of central similarity learning is used. The quantization loss is removed.
    \item \emph{CSMVH-quant}: The quantization loss is used. The BCE loss of central similarity learning is removed.  
    \item \emph{CSMVH-image}: Only the visual feature is used. The text embedding is removed.
    \item \emph{CSMVH-text}: Only the text embedding is used. The visual feature is removed.
    \item \emph{CSMVH-concat}: The image and text features are fused with concatenation. The multi-view fusion module is removed.
    \item \emph{CSMVH}: Our full framework.
\end{itemize}

The comparison results are presented in Table \ref{Tab:03}. Beginning with the loss function, when we solely use the quantization loss to train the proposed CSMVH, it can not learn anything just as we expected. The quantization loss can not perform any optimization on the embeddings. Because the CSMVH obtains data at random, all of the tasks have relatively poor mAP. On the contrary, the BCE loss of central similarity learning can help the CSMVH learn the embedding well. Due to the absence of a quantization constraint, CSMVH-central performs marginally worse than the entire framework.

Moving on to the multi-view features, CSMVH-text is barely superior to random selection (CSMVH-quant). In every task, CSMVH-image performs significantly better than CSMVH-text. It can be seen that the image features contain more useful information than text. Searching with images is more likely to find a related result. Our method already performs better with concatenated multi-view features than the state-of-the-art methods. But the multi-view fusion module takes it one step further, reaching even higher mAP. 

To conclude, the multi-view fusion module can improve the complementarity and representation capability of the fused feature. And the central similarity learning enhances the discriminative capability of hash codes. Both image and text features can provide rich information for the multi-view hashing task.

\subsection{Convergence Analysis}
\begin{figure}
	\centering
	\subfigure{\includegraphics[scale=0.35]{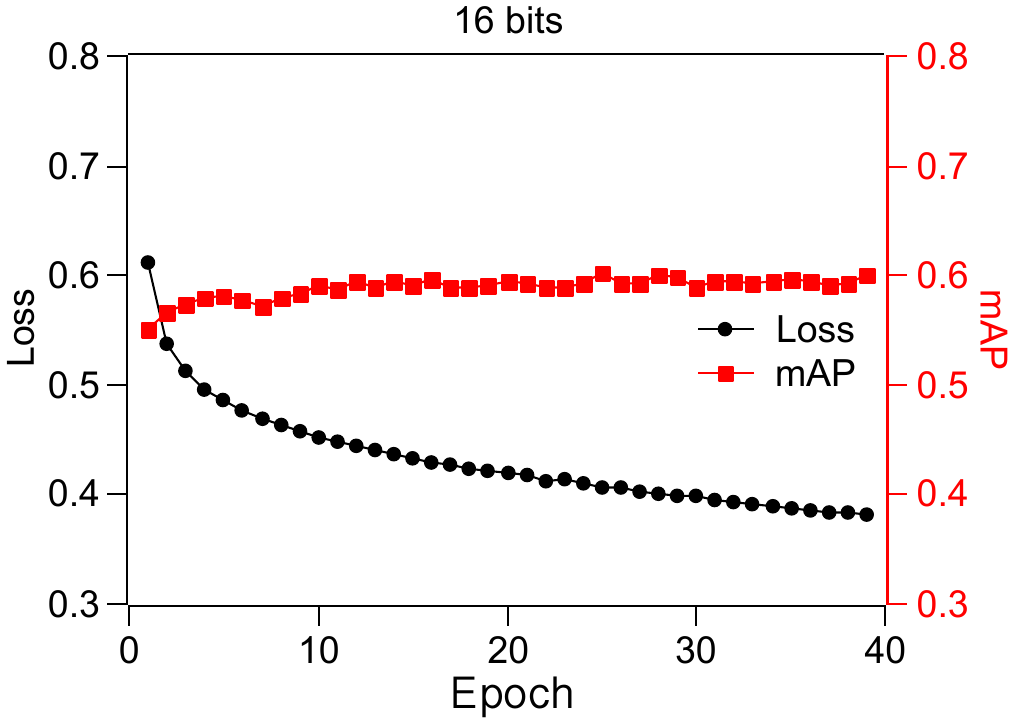}}
	\subfigure{\includegraphics[scale=0.35]{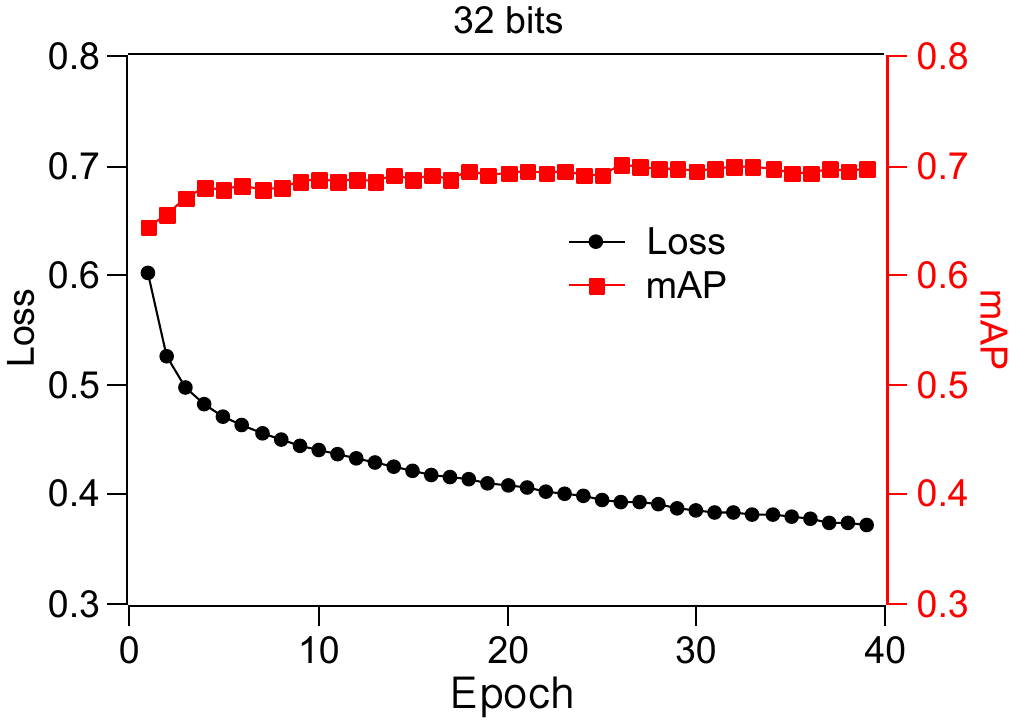}}
	\subfigure{\includegraphics[scale=0.35]{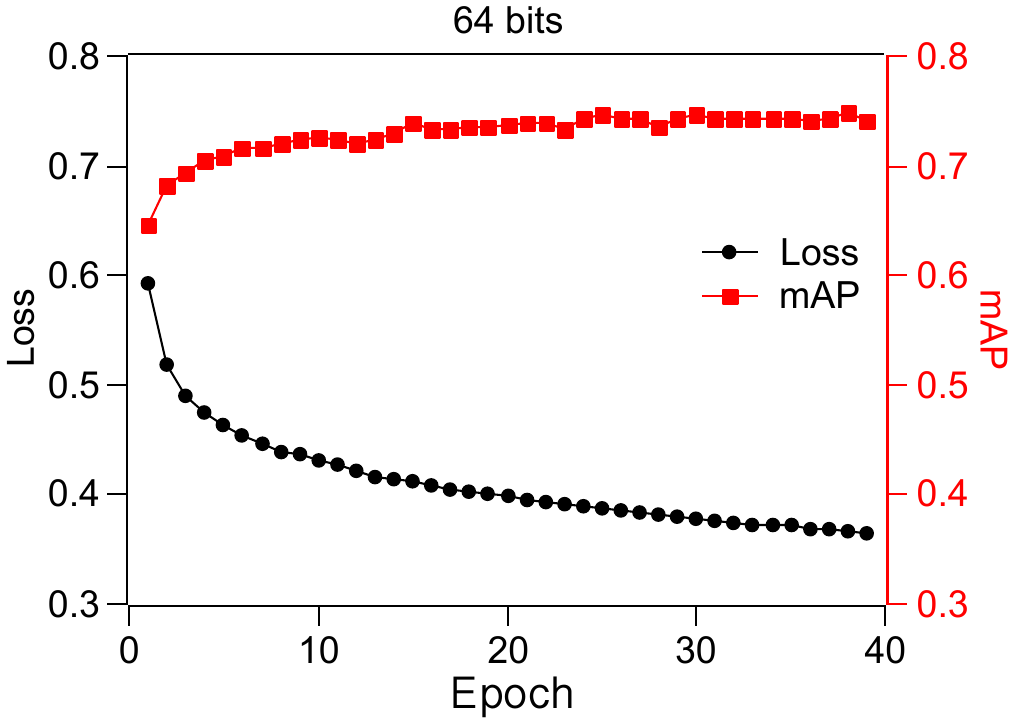}}
	\subfigure{\includegraphics[scale=0.35]{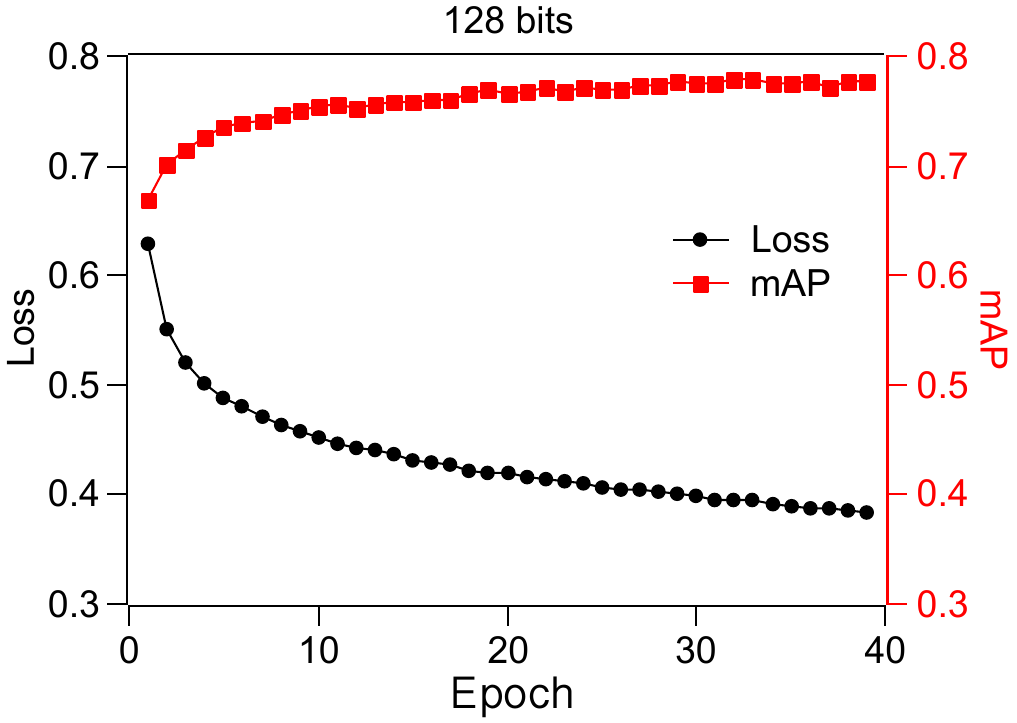}}
	\caption{The training loss curve and test mAP curve on MS COCO dataset. In this figure, the upper curve is the test mAP, and the bottom is the training loss. We notice that after around 40 epochs, our method converges, and the training loss no longer decreases. The test mAP increases rapidly during the first 10 epochs and slowly converges afterward. No overfitting is observed during the experiment.}
	\label{fig:03}
\end{figure}

We do tests to see how well the CSMVH can generalize and converge. Using the MS COCO dataset, we perform hash benchmarks using various code lengths. The results are shown in Fig. \ref{fig:03}. The graphic displays training loss and tests mAP. The loss gradually lessens as the training progresses. The loss stabilized at 40 epochs, indicating that the local minimum is attained. When the experiment starts, the mAP for the test metric quickly increases. The test mAP remains steady after 20 epochs. No worsening on the test MAP is seen after additional training, which suggests high generalization capacity and no overfitting. With other datasets, we notice a similar convergence outcome. The convergence of our method is promised on the popular datasets. 

\subsection{mAP@K and Recall@K}
\begin{figure}
	\centering
	\subfigure{\includegraphics[scale=0.35]{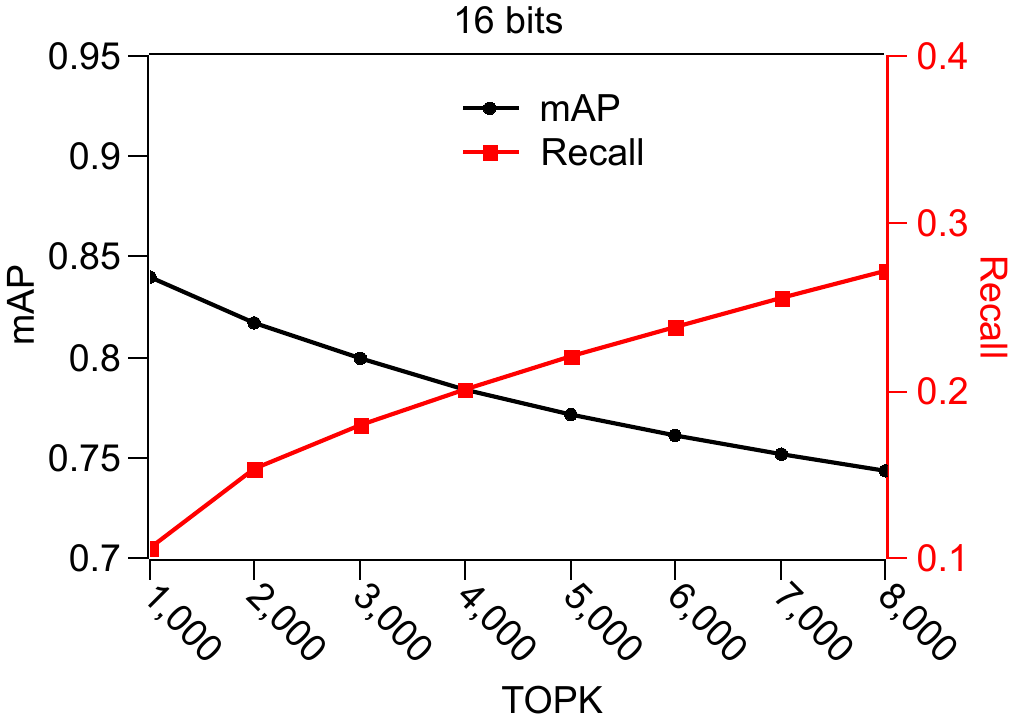}}
	\subfigure{\includegraphics[scale=0.35]{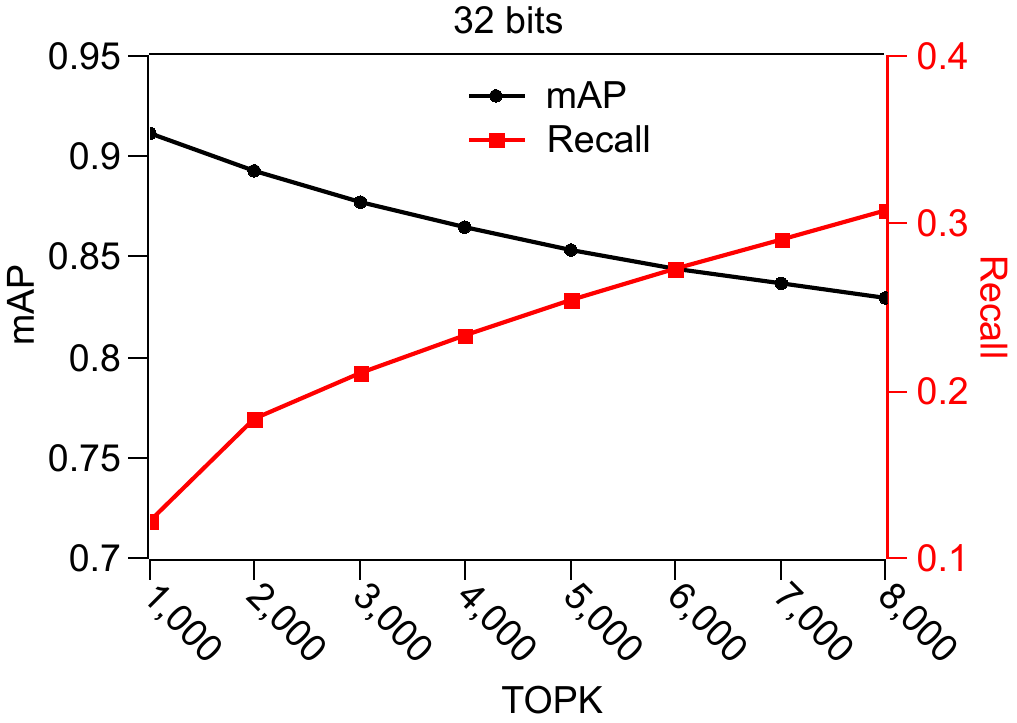}}
	\subfigure{\includegraphics[scale=0.35]{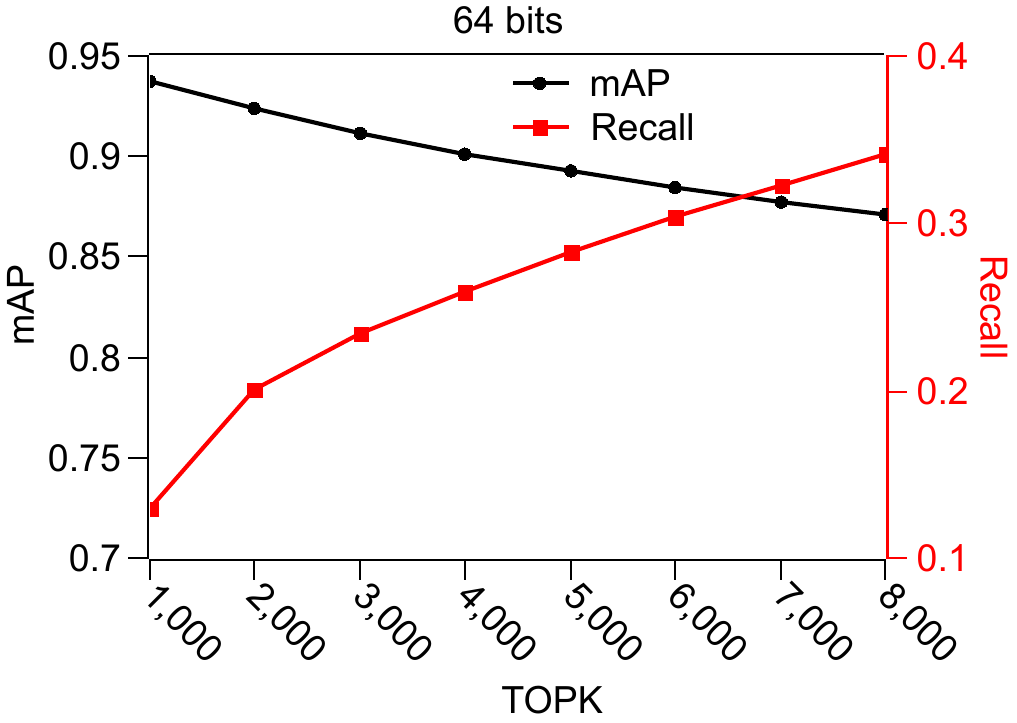}}
	\subfigure{\includegraphics[scale=0.35]{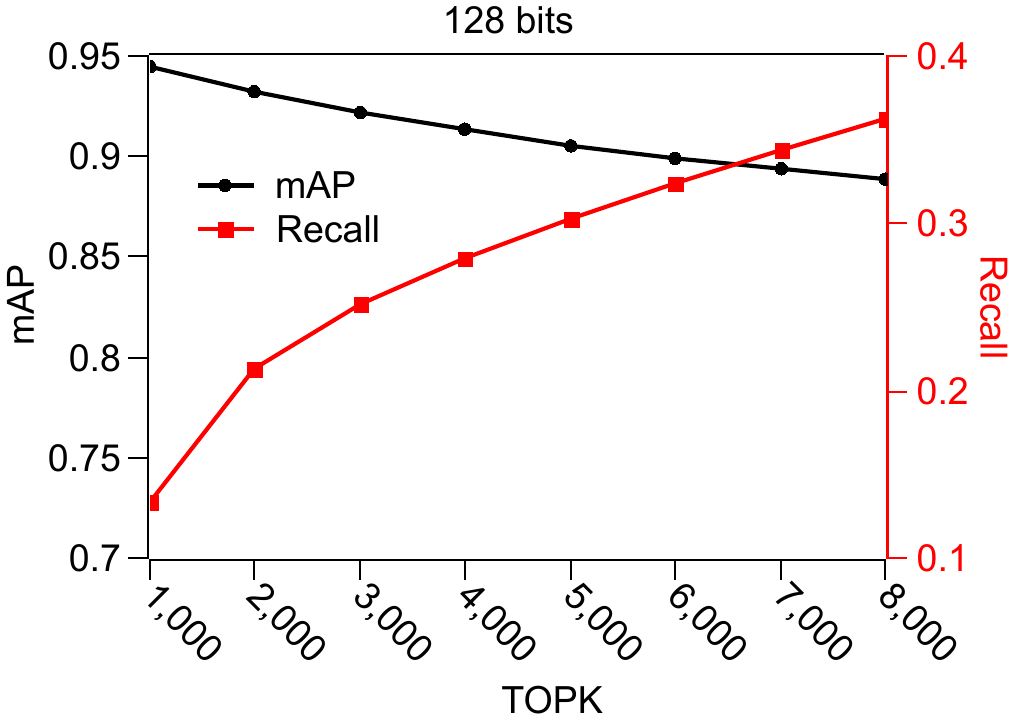}}
	\caption{The mAP@K and Recall@K curves on the MS COCO dataset. The recall curves show a nice linear increase as K increases, while the mAP only slightly decreases. We notice some sweet spots during the experiment. For example, in the 128-bit group, when $TOPK=5000$, the mAP keeps the same as $TOPK=4000$, but the recall keeps linearly increasing.}
	\label{fig:04}
\end{figure}

The matching mAP@K and Recall@K curves for the MS COCO dataset with increasing numbers of retrieved samples are shown in Fig. \ref{fig:04} for various code lengths. The recall curve exhibits quick linear growth, whereas the mAP of the four graphs somewhat declines as the $TOPK$ grows. The tendency demonstrates how well our method does in the retrieval tasks. The majority of people focus their attention on the first few results of the received data. In this case, the precision of our method is considerably higher. Compared to normal users, experts typically browse more results. With respect to the expansion of the retrieved data, our method can deliver a recall that grows linearly. When conducting their search, professionals may anticipate dependable, high-quality outcomes. The CSMVH can purposely give satisfying retrieval results for various user groups.

\section{Conclusion and Future Work}
We propose a new deep multi-view hashing method termed CSMVH. To address multi-view hashing issues, it makes use of central similarity learning. The CSMVH is proposed to conquer two main challenges of the multi-view hashing problem. In our experiment, CSMVH tends to yield consistent improvement in mAP with the growing length of hash code without any signs of performance degradation or overfitting. CSMVH is less computationally intensive than current methods. Under multiple experiment settings, it delivers an impressive performance increase over the state-of-the-art methods. Impressively, our method performs even better in more complex hashing tasks. We also find some problems during the process of our experiments. For instance, the performance improvement becomes less and less noticeable as the hash code length increases for some datasets. To further enhance the proposed method, we will work on these problems.
\section{Acknowledgment}
This work is supported in part by the Zhejiang provincial ``Ten Thousand Talents Program'' (2021R52007), the National Key R\&D Program of China (2022YFB450
\\
0405), and the Science and Technology Innovation 2030-Major Project (2021ZD01
\\
14300).

\bibliographystyle{splncs04}
\bibliography{main}

\end{document}